\documentclass{article}
\usepackage{ijcai19}

\usepackage{times}
\usepackage{soul}
\usepackage{url}
\usepackage[hidelinks]{hyperref}
\usepackage[utf8]{inputenc}
\usepackage[small]{caption}
\usepackage{graphicx}
\usepackage{subcaption}
\usepackage{amssymb}
\usepackage{amsmath}
\usepackage{booktabs}
\usepackage[linesnumbered,ruled,norelsize]{algorithm2e}
\usepackage{makecell}

\makeatletter
\newcommand{\removelatexerror}{\let\@latex@error\@gobble}
\makeatother

\newcommand{\x}{\mathbf{x}}
\newcommand{\y}{\mathbf{y}}
\newcommand{\yh}{\mathbf{\hat{y}}}

\newcommand{\ba}{\mathbf{a}}
\newcommand{\e}{\mathbf{e}}
\newcommand{\E}{\mathbf{E}}
\newcommand{\bt}{\boldsymbol{\theta}}

\title{Advocacy Learning: \\ Learning through Competition and Class-Conditional Representations}

\author{
Ian Fox \And Jenna Wiens
\affiliations
Department of Computer Science and Engineering, University of Michigan, Ann Arbor, USA
\emails
\{ifox, wiensj\}@umich.edu
}

\begin{document}

\maketitle
\begin{abstract}
   We introduce \textit{advocacy learning}, a novel supervised training scheme for attention-based classification problems. Advocacy learning relies on a framework consisting of two connected networks: 1) $N$ \textit{Advocates} (one for each class), each of which outputs an argument in the form of an attention map over the input, and 2) a \textit{Judge}, which predicts the class label based on these arguments. Each Advocate produces a class-conditional representation with the goal of convincing the Judge that the input example belongs to their class, even when the input belongs to a different class.  Applied to several different classification tasks, we show that advocacy learning can lead to small improvements in classification accuracy over an identical supervised baseline. Though a series of follow-up experiments, we analyze when and how such class-conditional representations improve discriminative performance. Though somewhat counter-intuitive, a framework in which subnetworks are trained to competitively provide evidence in support of their class shows promise, in many cases performing on par with standard learning approaches. This provides a foundation for further exploration into competition and class-conditional representations in supervised learning.
\end{abstract}

\section{Introduction}
\begin{figure}[!ht]
	\centering
	\begin{subfigure}{0.25\linewidth}
		\includegraphics[width=\textwidth]{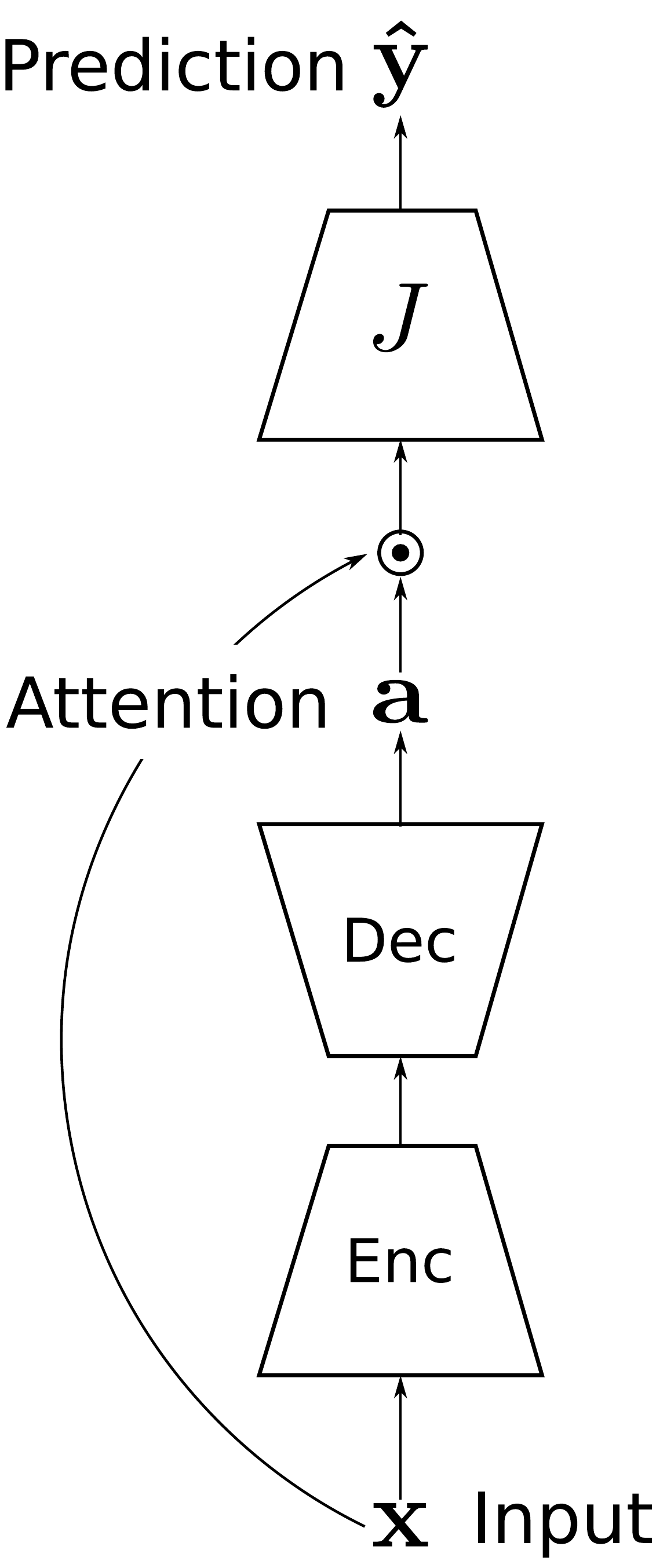}
		\caption{}
		\label{fig:attn}
	\end{subfigure}
	~ 
	\begin{subfigure}{0.5\linewidth}
		\includegraphics[width=\textwidth]{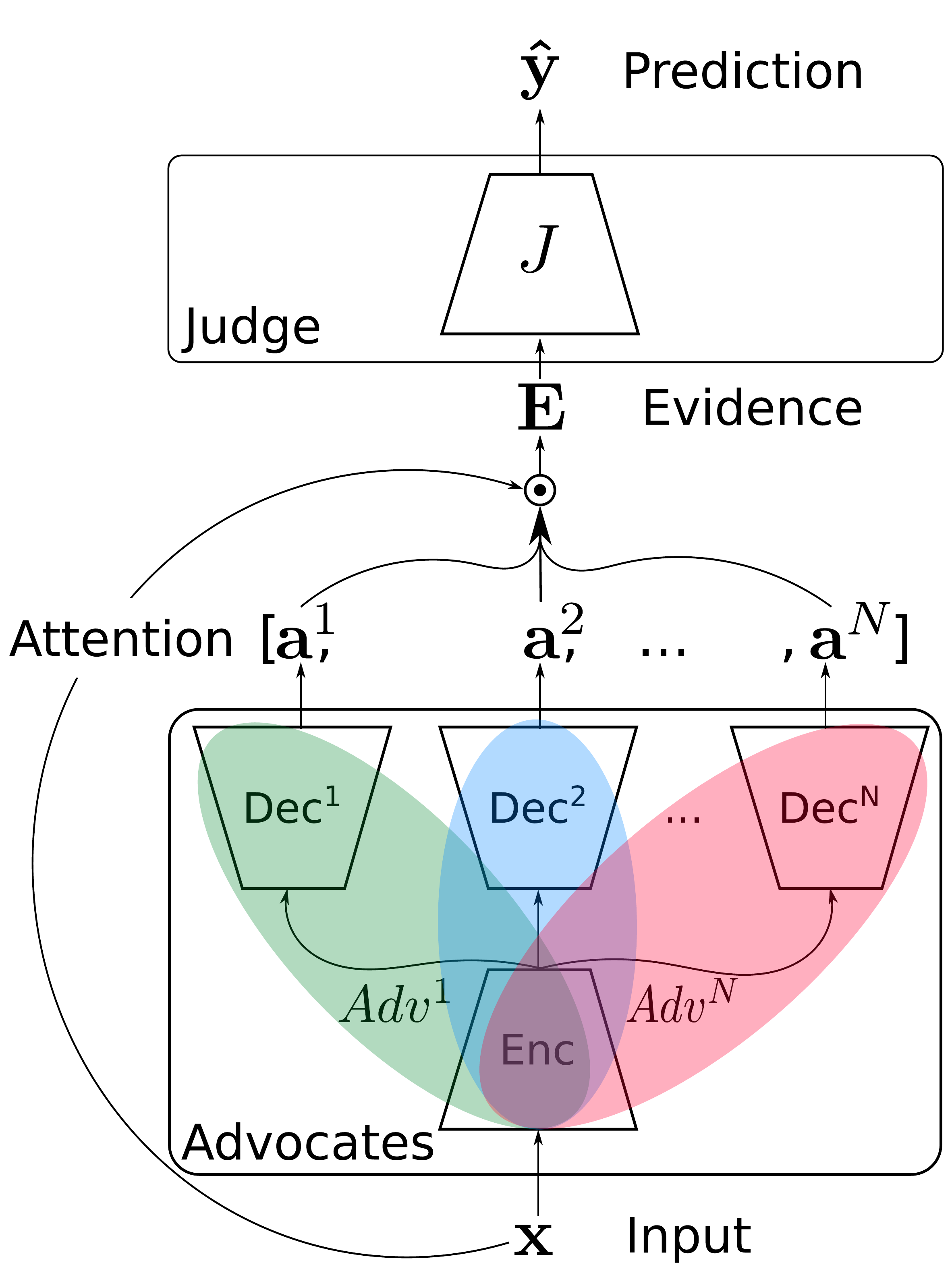}
		\caption{}
		\label{fig:advocate}
	\end{subfigure}
	~
	\begin{subfigure}{0.32\textwidth}
		\includegraphics[width=\textwidth]{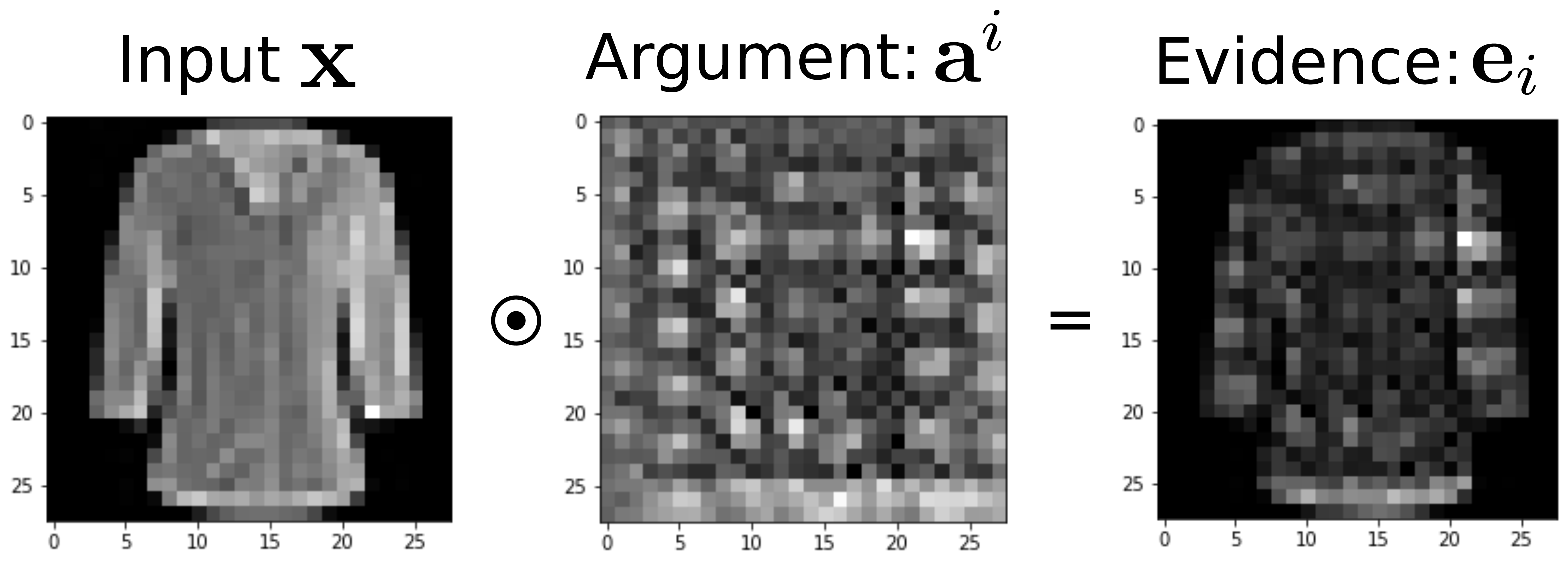}
		\caption{}
		\label{fig:saliency}
	\end{subfigure}
	\caption{\textbf{a)} A simple single-attention framework. The encoder-decoder produces an attention map $\mathbf{a} \in \mathcal{R}^{n \times n}$, which is combined with the input $\mathbf{x} \in \mathcal{R}^{n \times n}$ via an element-wise product (indicated by $\odot$) to create the input to the decision module, or Judge $J$. \textbf{b)} Our advocacy learning framework. Each decoder $Dec^i$ is trained separately to output a class-conditional attention map, or argument $\mathbf{a}^i \in \mathcal{R}^{n \times n}$, which is combined with the input via element-wise product (separately for each attention map) to create evidence $\mathbf{E} = [\mathbf{e}_0, \dots, \mathbf{e}_N]$, where $\mathbf{e}_i = \mathbf{a}^i \odot \mathbf{x}$ is evidence supporting class $i$. Each advocate is shown in a different color, the number of Advocates is equal to the number of classes. \textbf{c)} An example of an attention map $\mathbf{a}^i$ used to generate evidence $\mathbf{e}_i$.}\label{fig:framework}
\end{figure}

In recent years, researchers have proposed a large number of modifications to the standard supervised learning setting with the goal of improving performance \cite{parascandolo_learning_2018,vaswani_attention_2017}. These modifications focus on training different parts of the network (\textit{i.e.}, subnetworks) to \textit{cooperate}. However, in several real-world settings, such as in economics and law, agents who \textit{compete} are critical for identifying good solutions. While recent work in adversarial networks investigates the use of competition for training models, the final model evaluated (\textit{i.e.}, the generator) is cooperative \cite{goodfellow_generative_2014}. In contrast, we investigate a training scheme in which subnetworks compete during training \textit{and} evaluation. In our model, subnetworks compete to provide evidence in the form of class-conditional attention maps. Here, we use the term `attention map’ to refer to a filter that indicates parts of the input that are useful for accurate classification, similar to the idea of saliency \cite{itti_model_1998}. An example of a standard network architecture with attention is given in \textbf{Figure \ref{fig:attn}}. We hypothesize that class-conditional attention maps could offer advantages over standard attention maps by emphasizing portions of the input indicative of their class.

Our proposed approach consists of two main subnetworks: a single Judge and multiple Advocates (see \textbf{Figure \ref{fig:advocate}}). Each Advocate produces an attention map that \textit{advocates} for a particular class. A decision is reached by the Judge, which weighs the arguments produced by the Advocates. For this approach to work well, there must be a balance among the Advocates (so that each Advocate can influence the Judge), and the Judge must be able to effectively use the given evidence (so as not to be deceived by the advocates). We achieve this balance via \textit{advocacy learning}, which trains the components jointly, but according to multiple objectives. These different objectives are key to striking the right balance between providing strong but factual evidence. We also explore a variant, honest advocacy learning, where the Advocates are not trained to deceptively compete with one another, but still provide class-conditional attention maps. In a series of experiments, we compare advocacy learning to several baselines in which the entire network is trained according to the same standard objective. Across several image datasets, we observe a small but consistent improvement in classification accuracy by using class-conditional attention maps.

\section{Methods}
We propose a novel approach to optimizing networks for supervised classification that encourages class-conditional representations of evidence in the form of attention maps. We hypothesize that, depending on how they are learned, class-conditional attention maps could offer advantages over standard attention maps by encouraging competition among components of the network. At a high level, the Judge learns to solve the classification problem given some evidence, while each Advocate supplies that evidence by arguing in support of their class. This setup disentangles evidence supporting each class, encouraging strong class-conditional representations. Advocacy learning consists of both a specific architecture (\textit{i.e.}, Advocate and Judge subnetworks) and a specific method for training, both are described below.

\subsection{Problem Setting}
We consider the task of solving a multi-class classification problem in a supervised learning setting with element-wise attention. We assume access to a labeled training set consisting of labeled examples $\{\x,y\}$, where $\x \in \mathbb{R}^{d}$ (where $d$ may be a product $d_1 \times d_2$, such as in an image) and $y \in \{1, \dots, N\}$, where $N$ is the number of classes. We refer to the one-hot label distribution entailed by $y$ as $\y$, so $\y[y] = 1$ and $\y[j] = 0$ for all $j \ne y$. We use square brackets for indexing into a vector. We focus on deep learning methods due to their suitability for representation learning. We indicate the parameters of deep models as $\bt$ and subscript them according to the particular subnetwork being referenced. The parameters of the Judge are $\bt_J$ and the parameters for each Advocate $i$ are $\bt_i$ where $i \in \{1, \dots, N\}$. Our proposed approach aims to solve the multi-class classification problem through a novel training scheme, designed to produce class-specific evidence. 

\subsection{Network Architecture}
As mentioned above, our proposed approach is composed of two sets of modules: multiple Advocates and a single Judge. A high-level overview of our architecture, which we call an Advocacy Net, is given in \textbf{Figure \ref{fig:advocate}}. Here, for context, we describe a generic framework, providing specific implementation details in a later section.

\subsubsection{Advocate Subnetwork}
This subnetwork consists of $N$ Advocates $Adv^i$, where $i$ corresponds to the index of the class the Advocate represents. Given the input $\x$, each advocate generates an argument in the form of an attention map $Adv^i(\x; \bt_i) \rightarrow \ba^i \in [0, 1]^{d}$. The Advocate modules produce an attention map with dimensionality equal to the input using a convolutional encoder-decoder, as is standard for producing pixel-level output in images \cite{badrinarayanan_segnet:_2017}. Note that for complex input, such as medical images, other fully convolutional architectures such as U-Nets may be more appropriate \cite{ronneberger_u-net:_2015}. Based on these attention maps, each Advocate presents an argument $\e_i$ (or evidence) to the Judge in the form of an element-wise product between attention maps and the input, $\e_i = \ba^i \odot \x$.

Each Advocate is trained to emphasize aspects of the input indicative of the Advocate's class. This differs from a supervised attention map, which focuses on aspects of the input indicative of the inputs underlying class. In our implementation, Advocates share some underlying evidence in the form of a shared encoder. This allows the Advocates to share useful representational abstractions.

\subsubsection{Judge Subnetwork}
The Judge $J$ takes as input the combined evidence $\E = [\e_1, \dots, \e_N] \in \mathbb{R}^{N \times d}$, and outputs a probability distribution over classes $J(\E; \bt_J) \rightarrow \yh$. We make specific class predictions using $\arg \max (\yh)$. The architecture of the Judge is flexible; the only limitation is that the input size must be proportional to the total number of classes. In our implementation, the Judge module is a convolutional network with fully connected output layers.

Though certain constraints on the architecture of the network are necessary, the interplay among the modules and how they are trained are key to the advocacy learning framework. Trained end-to-end with the objective of minimizing training loss, there would be no difference between the proposed architecture and a network with multiple attention channels. In the next section, we describe the key differences in how we train the Advocates vs. the Judge.

\subsection{Training Algorithm}
The complete advocacy learning algorithm is presented in \textbf{Algorithm 1}. We learn the parameters of the Judge subnetwork by minimizing the cross-entropy loss: $CE(\yh, y) = -\log \yh[y]$, as is standard for classification. The Advocates are trained according to a different objective. In particular, Advocate $i$ is trained by minimizing the advocate cross-entropy loss: $CE^{A}(\yh, i) = -\log \yh[i]$. Under this objective, each Advocate is trained to represent samples from all classes as its own. We also consider a variant, called honest advocacy learning, in which the Advocates are not trained to deceive, but aim to minimize: 
\[
CE^{HA}(\yh, i, y) = \
\begin{cases} 
-\log \yh[i], & \text{if } i = y \\
0, & \text{otherwise }
\end{cases}
\]

We optimize the parameters of the Judge and Advocates by making a prediction and then interleaving steps of gradient descent, updating the Judge and each Advocate individually according to their respective loss functions. At each step, we freeze the parameters of all other subnetworks. 

\begin{figure}[!t]
\removelatexerror
\begin{algorithm}[H]
	\SetKwInOut{Input}{Input}
	\SetKwInOut{Output}{Output}
	
	\Input{Labeled training data $\mathbf{D} = \{\x_k, y_k\}_{k=1}^S$ where $S$ is the number of samples, $\x_k \in \mathcal{R}^d$, and $y_k \in \{1, \dots N\}$}
	\Output{Trained Network $A = (J, Adv^1, \dots Adv^N)$}
	Initialize parameters for Judge $\bt_J$ and Advocates $\bt_1, \dots, \bt_N$\;
	\While{training}
	{   \tcc{Showing single sample for simplicity}
		Draw example $(\x, y) \sim \mathbf{D}$\;
		\For{$i \in \{1, \dots N\}$}
		{
			$\ba^i \gets Adv^i(\x; \bt_i)$\; 
			$\mathbf{e}_i = \ba^i \odot \x$\;
		}
		$\E \gets [\mathbf{e}_1, \dots, \mathbf{e}_N]$\;
		$\yh \gets J(\E; \bt_J)$\;
		$L_J = - \log(\yh[y])$ \tcp*[r]{Cross-Entropy Loss}
		$\bt_J \gets \bt_J - \eta \bigtriangledown_{\bt_J} L_J$\;
		\For{$i \in \{1, \dots N\}$}
		{
		    \If{not honest \textbf{ or } $i = y$}{
		        \tcc{Honest Advocates update on true examples}
		        $L_{Adv^i} = -\log(\yh[i])$\;
		        $\bt_i \gets \bt_i - \eta \bigtriangledown_{\bt_i} L_{Adv^i}$\;
		    }
		}	
	}
	return $A = (J, Adv^1, \dots, Adv^N)$
	\caption{Advocacy Learning Algorithm}
\end{algorithm}
\end{figure}

\section{Baselines \& Experimental Setup}
We evaluate our proposed advocacy learning approach across a variety of tasks and compare against a series of different baselines. In this section, we explain each baseline and provide implementation details.

\subsection{Model and Baselines}
On all datasets, we compare (honest) advocacy learning against two baselines that incorporate attention:
\begin{itemize}
 \setlength\itemsep{0em}
	\item \textbf{Attention Net:} this baseline modifies the Advocacy Net architecture by removing all but one attention module. The one remaining module is trained using a standard end-to-end optimization approach minimizing cross entropy loss. This allows us to compare advocacy learning against a similar model using standard supervision.
	\item \textbf{Multi-Attention Net:} two differences exist between Attention Nets and Advocacy Nets: the optimization procedure and the architecture. To tease apart these two aspects, we include a comparison against a model with an identical architecture, but trained using a standard end-to-end loss. 
\end{itemize} 

\subsection{Implementation Details}
We implement our models using PyTorch \cite{paszke_automatic_2017}. Our specific model architecture (number of layers, filters, \textit{etc.}) is available via our public code release\footnote{\url{https://github.com/igfox/advocacy-learning}}.

In our experiments, we optimize the network weights using Adam \cite{kingma_adam:_2014} with a learning rate of $10^{-4}$, and use Dropout and batch normalization to prevent overfitting. We split off 10\% of our training data to use as a validation set for early stopping. We cease training when validation loss fails to improve over 10 epochs. Model performance is reported on the canonical test splits for each dataset. We regularize the attention maps by adding a penalty proportional to the L1-norm of the map to encourage sparsity consistent with common notions of attention. Parameters were initialized using the default PyTorch method.

\section{Results and Discussion}
We begin by examining the performance of advocacy learning across three image datasets, analyzing when and how advocacy learning impacts performance. We then present performance on a significant real-world medical dataset and modified datasets designed to highlight the effects of competition and deception in learning. We conclude by providing some intuition for why advocacy learning works.

\subsection{Advocacy Learning on Multi-Class Balanced Image Data}
We begin by examining the performance of our Advocacy Net variants and baselines on two publicly available image classification datasets: MNIST and Fashion-MNIST \cite{xiao_fashion-mnist:_2017}. As described above, the parameters of the Advocate modules are not optimized to improve overall classification performance, and as a result their inclusion could lead to a reduction in performance (by deceiving the Judge). However, we find that this is not the case (\textbf{Table \ref{table:judge}}). On these two datasets, advocacy learning does as well as or outperforms all baselines. The improvement is most pronounced in Fashion-MNIST, perhaps due to the denser images or the baseline performance allowing room for improvement. Moreover, this difference is not solely attributable to the class-conditional nature of the attention-maps; across both datasets the Advocacy Nets outperform the Honest Advocacy Nets (test accuracy $99.42$ \textit{vs.} $99.32$ in MNIST and $91.62$ \textit{vs.} $90.81$ in FMNIST). This suggests that deception, in addition to competition, can help produce useful attention maps.

We also examined performance on a more challenging image classification problem, CIFAR-10. When using a similar architecture and training procedure as in the above datasets, we found the Advocacy Net improved over other approaches (in particular, test accuracy $83.47$ \textit{vs.} $79.73$ for the Multi-Attention Net), though both approaches performed poorly relative to state-of-the-art. Thus, we examined replacing the Judge network with a ResNet-110, using a fixed-epoch training scheme with learning rate decay (similar to a popular open source implementation\footnote{\url{https://github.com/bearpaw/pytorch-classification}}). This increased the performance of the Multi-Attention Net and drastically lowered the Advocacy Net performance (test accuracy $30.54$ \textit{vs.} $92.01$), suggesting that under our current training scheme Advocacy Learning is unstable when the Judge is high-capacity relative to the Advocates. Notably, Honest Advocacy Learning continues to perform well, beating the Multi-Attention Net (test accuracy $92.68$ \textit{vs.} $92.01$).

\begin{table}
	\centering
	\resizebox{\columnwidth}{!}{
	\scalebox{0.8}{
	\begin{tabular}{lll}
		\toprule
		& \multicolumn{2}{c}{Dataset} \\ \cmidrule(r){2-3}
		Model    & MNIST & FMNIST \\
		\midrule
		Attn Net & $99.16 \pm 0.30$ (0\%) & $89.71 \pm 0.86$ (0\%)\\
		\makecell[l]{Multi-Attn  Net} & $99.33 \pm 0.09$ (20\%)& $90.11 \pm 0.40$ (4\%) \\
		\midrule
		\makecell[l]{Honest Adv. Net} & $99.32 \pm 0.08$ (19\%) & $90.81 \pm 0.34$ (11\%) \\
		Advocacy Net & $\mathbf{99.42 \pm 0.05}$ (31\%)& $\mathbf{91.62 \pm 0.41}$ (19\%) \\
		\bottomrule
	\end{tabular}
	}
	}
	\caption{Accuracy $\pm$ standard deviation over 5 random seeds on the datasets among the various models. Values in parentheses show reduction in error rate relative to Attention Net. Note that the difference between the Advocacy Net and other approaches are consistent across the individual seeds.}
	\label{table:judge}
\end{table}

\subsection{Impact of Class Conditional Attention}
Results in \textbf{Table \ref{table:judge}} demonstrate that class-conditional attention can improve upon supervised attention maps. Honest Advocacy Nets perform similarly to Multi-Attention Nets on MNIST and slightly better on FMNIST ($99.32$ \textit{vs.} $99.33$ and $90.81$ \textit{vs.} $90.11$). These architectures differ only in that the attention maps in the Honest Advocacy Net are class-conditional. The competition introduced by the Advocacy Net further improves performance, outperforming the Multi-Attention Net on both datasets. 

\begin{figure}[t]
	\centering
	\includegraphics[width=\columnwidth]{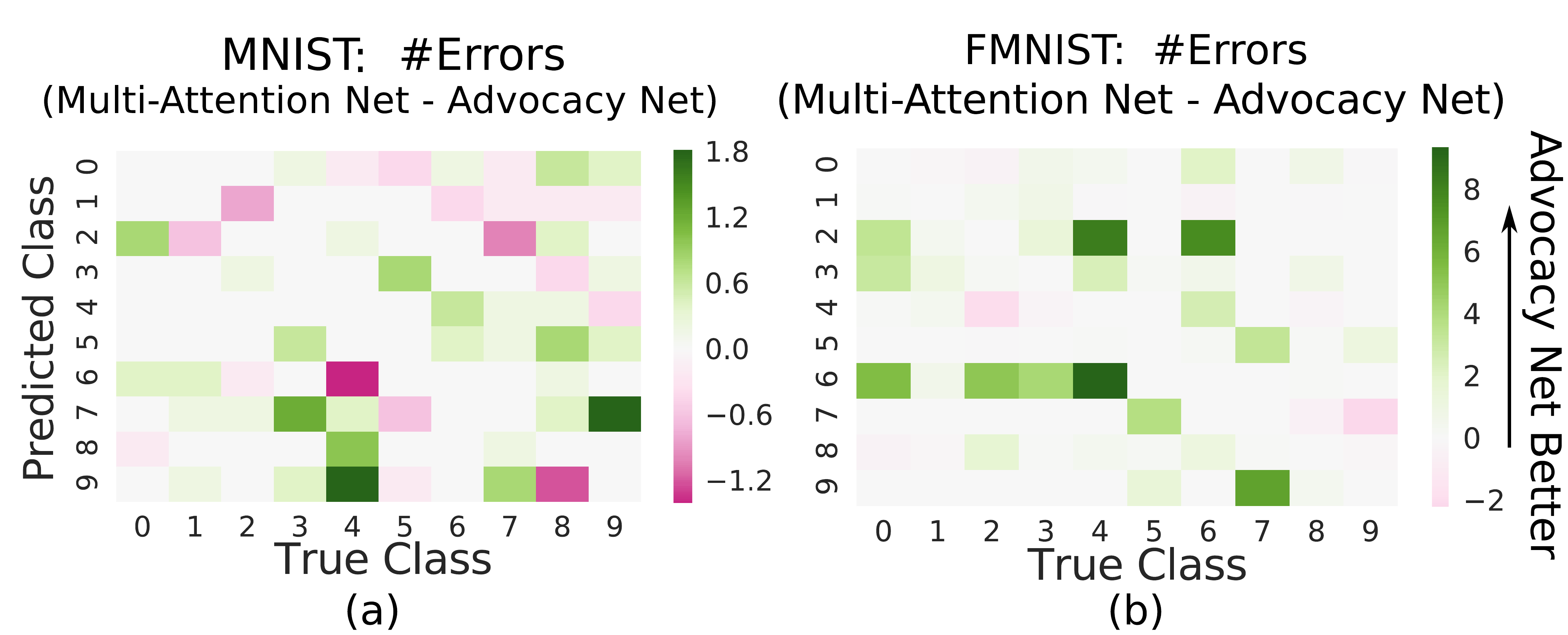}
	\caption{Averaged difference across five runs in confusion matrices between the multi-attention and Advocacy networks ($\ge 0$ means the Advocacy net performed better) on a) MNIST and b) FMNIST. We zero out the diagonal elements to focus on misclassification. A positive number means the Multi-Attention Net made more misclassifications than the Advocacy Net and vice-versa. We observe the Advocacy Net tends to improve performance across classes, but can make certain morphologically similar examples (\textit{i.e.}, 8 vs 9 in \textbf{a}) more difficult.}
	\label{fig:confusion}
\end{figure}

To further compare advocacy learning with the end-to-end supervised baselines, we plot the averaged difference between the confusion matrices of the Multi-Attention Nets and Advocacy Nets (\textbf{Figure \ref{fig:confusion}}). Overall, we observe that Advocacy Nets result in improvements for a subset of class pairs (\textit{e.g.}, classes 4 and 9), but leave the majority of predictions unchanged. On MNIST (\textbf{Figure 2(a)}), we find a few examples where advocacy learning lowers performance. In particular, the Advocacy Net is more likely to misclassify 8s as 9s; though the reverse error rate (9s as 8s) does not increase. This is likely due to the asymmetric morphological relationship among the digits: using per-pixel attentions in $[0, 1]$, an 8 can be obscured to look like a 9, but the converse is less likely. The Advocacy Net appears to help emphasize curves in the input (reducing the instances with 9 classified as 4 or 7 classified as 9). On Fashion-MNIST (\textbf{Figure 2(b)}) we observe that certain class pairs (pullovers or coats \textit{vs.} shirts) are markedly improved, while most others are unaffected. This suggests that advocacy learning improves performance by distinguishing among classes with similar morphology; though this analysis is confounded by the fact that it tends to be those class pairs that have the greatest potential room for improvement.


\begin{figure*}
\centering
\includegraphics[width=1.5\columnwidth]{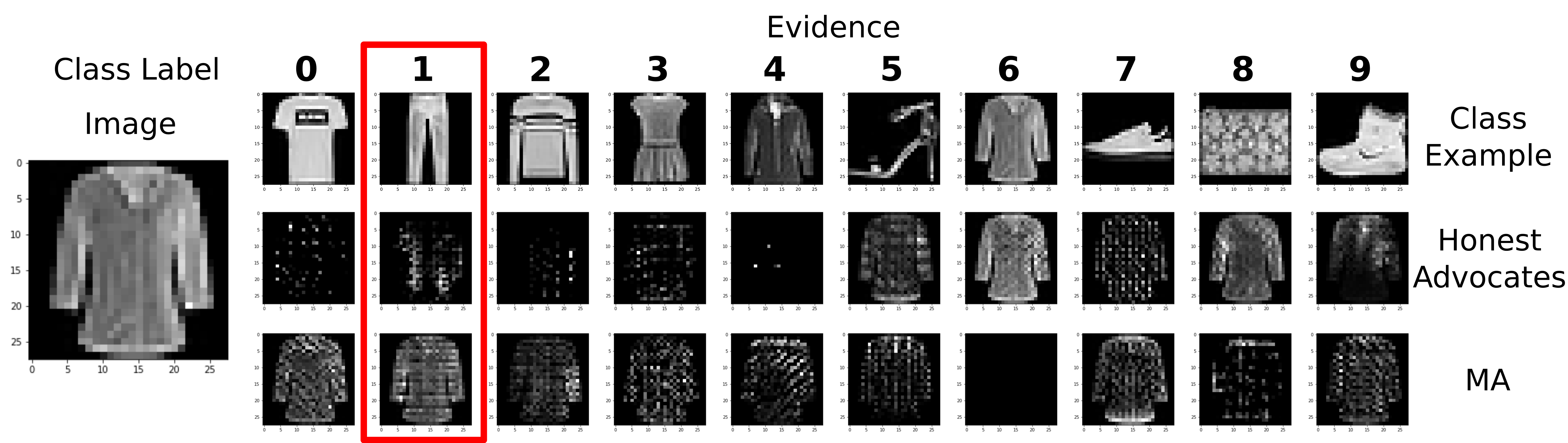}
\caption{Evidence generated from a Fashion-MNIST example. The top row shows a sample from the class the column represents. The second row shows evidence from an Honest Advocate Net (the order corresponds to class). The bottom row shows evidence generated by the Multi-Attention Net (the ordering is arbitrary). The image is an example of class 6 (shirts). Of particular note is the argument generated by the Advocate for class 1 (pants).}
\label{fig:qualitative_fmnist}
\end{figure*}

\begin{table*}[ht!]
	\centering
	{
	\scalebox{0.95}{
	\begin{tabular}{lllll}  
	\toprule
		& \multicolumn{3}{c}{Dataset (Metric)} \\ \cmidrule(r){3-5}
		& \multicolumn{2}{c}{MIMIC} & \multicolumn{2}{c}{MNIST (Accuracy)} \\
		Model & (AUROC) & (AUPR) & Imbalanced & Binary \\
		\midrule
		Attention Net & $\mathbf{83.29 \pm 0.84}$ (0\%) & $45.79 \pm 1.80$ (0\%) & $98.68 \pm 0.48$ (0\%) & $99.23 \pm 0.22$ (0\%) \\
		Multi-Attention Net & $82.80 \pm 0.94$ (-3\%) & $45.74 \pm 1.71$ (0\%) & $99.00 \pm 0.13$ (24\%) & $\mathbf{99.32 \pm 0.14}$ (11\%) \\
		\midrule
		Honest Advocate Net & $\mathbf{83.22 \pm 0.93}$ (0\%) & $\mathbf{46.34 \pm 1.73}$ (1\%) & $\mathbf{99.17 \pm 0.06}$ (37\%) & $\mathbf{99.31 \pm 0.13}$ (10\%) \\
		Advocate Net & $77.73 \pm 1.43$ (-33\%) & $39.03 \pm 4.58$ (-12\%) & $\mathbf{99.17 \pm 0.14}$ (37\%) & $98.72 \pm 0.58$ (-66\%) \\
		\bottomrule
	\end{tabular}
	}
	}
	\caption{Accuracy $\pm$ standard deviation over 5 random seeds on the modified MNIST datasets among the various models. The MIMIC results are reported in terms of AUPR. Values in parentheses show reduction in error rate relative to Attention Net.}
	\label{table:new_judge}
\end{table*}

Qualitative examples of attention maps from the Honest Advocates and Multi-Attention Net are given in \textbf{Figure \ref{fig:qualitative_fmnist}}. We found Honest Advocacy Nets gave denser (and thus more interpretable) attention maps than Advocacy Nets. We observe that both Advocacy Nets and Multi-Attention Nets generate a variety of attention maps with checkering characteristic of deconvolutional layers \cite{odena2016deconvolution}. An interesting example of class-conditional behavior is shown by the advocate for class 1. This Advocate, representing the ``pants" class, emphasizes the sides of the shirt, similar to pants legs.

\subsection{Balancing the Advocates and Judge}

Since the Advocates are encouraged to deceive the Judge, a natural question arises - How does the relative capacity of these components impact performance? To answer this question, we consider a variation of our Advocacy Net, in which we can vary the capacity of different pieces. We replace the convolutional layers in the Advocate encoder and Judge with some number of convolutional residual blocks, as in \cite{he_deep_2015}. By changing the number of blocks, we can increase/decrease the capacity of the Advocate or Judge. Architectural details of this modification can be found in our code. We varied the number of residual blocks from 1-5 and 1-3 in the Judge and Advocates respectively. We observed that the Advocacy Nets achieved the highest classification accuracy when Judge capacity was high and Advocate capacity was low (best accuracy $99.46$). We performed an identical architecture search with the Multi-Attention Net; there was no capacity setting that beat the best results attained by the Advocacy Net (best $99.34\%$ \textit{vs.} $99.46\%$). This suggests that it is important for the Judge to have more capacity than the Advocates; though our experiments on CIFAR suggest that a balance must be struck. While a high-capacity Judge is better able to use evidence provided by the Advocates, it may train slowly, and thus be more susceptible to deceptive Advocates.

To better understand the impact of deception in advocacy learning, we took a fully trained network and examined the effect of freezing the Judge while continuing to update the Advocates for both advocate and Honest Advocacy Nets. We found that training without the Judge did not affect network performance in the Honest Advocate Net, but decreased performance in the advocate net by $85\%$. Thus, adaptations by the Judge play a crucial role in maintaining advocate net accuracy.

Up to this point, we have considered only Advocacy Nets in which all Advocates share an encoder. Such an architecture could encourage implicit sharing of information, possibly tempering the negative effects of deception. To test this hypothesis, we evaluated an Advocacy Net without a shared encoder on MNIST and FMNIST. On both datasets (averaged across 5 runs), the Advocacy Net without a shared encoder achieved lower performance both in absolute terms and relative to an Honest Advocacy Net without a shared encoder ($98.29$ \textit{vs.} $99.05$ for MNIST, $86.47$ \textit{vs.} $89.29$ for FMIST). This suggests the shared encoder is an important way for Advocates to balance performance.

\subsection{Generalization to other Settings}

The results presented so far all involve multi-class image datasets with balanced classes. To explore how these assumptions affect the performance of advocacy learning, we applied advocacy learning to a large electronic health record (EHR) dataset, MIMIC III \cite{johnson_mimic-iii_2016}. This dataset, a publicly available repository of EHR data, has become an important benchmark in the machine learning for health community \cite{harutyunyan_multitask_2017}, and is helping to drive advances in precision health \cite{desautels_prediction_2016,maslove_path_2017}. We used the clinical time-series subset of the database for mortality prediction, as in \cite{harutyunyan_multitask_2017}. We also considered variants of MNIST that break the multi-class and balanced assumptions. These additional experiments test the generalizability of our findings to i) different data types (time series as opposed to images), ii) imbalanced classes, and iii) binary labels. Our results are presented in \textbf{Table \ref{table:new_judge}}. 

We report our results on MIMIC in terms of the area under the precision recall curve (AUPR) and the receiver operating characteristic curve (AUROC), since the task is binary with considerable class imbalance in the test set ($13.23\%$ positive). Notably, in this task advocacy learning performs slightly worse than all baselines. However, honest advocacy learning provides a small benefit relative to the baselines in terms of AUPR. 

This reversal of the results from \textbf{Table \ref{table:judge}} is interesting, and helps illuminate cases where advocacy learning may or may not work. There are many differences between MIMIC and MNIST/FMNIST that could explain why advocacy learning applied to MIMIC fails. First, this task involves classifying patients based on a feature vector of vital sign measurements. Our attention mask allows for input values to be reduced to some proportion between 0 and 1. This type of attention can allow for pictures of shirts to look like pants, or an 8 to look like a 9, but it may not allow for vitals from sick patients to look like vitals from healthy patients (or vice-versa). Other major differences, besides the data type, include class imbalance and the number of classes. As these are qualities that can be imbued in other datasets, we further examine them by modifying MNIST to create new datasets we call Imbalanced MNIST and Binary MNIST. 

In Imbalanced MNIST, we subsampled the training set, introducing class imbalance. After subsampling, the least represented class, 0, had 600 training samples, and each successive class had 600 additional samples. The test set remained unchanged. We found that class imbalance in the training set lowered the performance of all models by 0.1-0.3\% relative to those same models performance with balanced training data; the Advocacy Net was more strongly affected than the Honest Advocacy Net. However, both models resulted in similar accuracy (advocacy learning $99.17 \pm 0.14$ \textit{vs.} honest advocacy learning $99.17 \pm 0.06$). 

Binary MNIST contains only two classes: 4 and 9. The per-class number of examples in the training and test set were unchanged. We found that the switch to a binary formulation \textit{reduced} absolute performance for the Advocacy Net by $0.47\%$ relative to the performance of the full network evaluated only on 4s and 9s (99.19 \textit{vs.} 98.72), whereas the Honest Advocacy Net \textit{improved} binary performance by $0.32\%$. This decrease suggests that, in practice, the competition among many Advocates helps the Judge achieve good performance in the presence of deception.

In all datasets, the class-conditional attention provided by honest advocacy learning did not hurt, and in the presence of imbalanced data helped. This suggests the value of class-conditional representations, with or without deception.

\section{Related Work}

The fact that advocacy learning, which encourages deceitful subnetworks, works at all, let alone better than the baselines in some tasks, may be surprising. Several recent works, however, have found competition useful for learning \cite{goodfellow_generative_2014,silver_mastering_2017,sabour_dynamic_2017}. Specifically, adversarial relationships, or situations where subnetworks compete with one another, have recently garnered interest \cite{goodfellow_generative_2014,ghosh_multi-agent_2017}. For example, Ghosh \textit{et al.} examine a multi-generator setting, where each generator is encouraged to capture distinct portions of the class distribution. This resembles how advocates captures relevant evidence for their corresponding class. However, while the relationship between an Advocate and the Judge is not entirely cooperative (the Advocate may argue for an untrue class), neither is it entirely adversarial (the Advocate may argue for a true class). While these systems used competition among networks during training, competition within a network has been used as well. A winner-take-all competitive framework was found to lead to superior semi-supervised image classification performance \cite{makhzani_winner-take-all_2015}, and the dynamic routing used in Capsule Networks can been seen as type of competition \cite{sabour_dynamic_2017}. 

Beyond this work in competition, advocacy learning is related to both i) input transformation and ii) debate agents. First, others have considered transformations of the input in the context of improving classification. Parascandolo \textit{et al.} proposed the use of a mixture-of-experts model to learn inverse data transforms to improve performance. The competition among experts to provide quality transforms of the input resemble the competition among Advocates to convince the Judge. Our work differs in that: 1) our Advocate updates are unsupervised, 2) the goal of the Advocates is not to improve performance, and 3) Advocates are assigned classes. Hong \textit{et al.} examined the use of class-conditional attention for improving semantic segmentation. Their attention maps are similar to the argument maps generated by Advocacy Nets, but are trained with an end-to-end objective. Similarly, Capsule Nets \cite{sabour_dynamic_2017} use class-specific capsules at the output layer to define a class-conditional parse of the input. This can be viewed as a bottom-up analog to the class-specific attention maps produced by Advocates, but it trained end-to-end unlike advocacy learning.

Second, work in AI debate sets up a similar task to our own, training agents to `debate' in order to convince a Judge about the class associated with an input \cite{irving_ai_2018}. The authors use Monte-Carlo tree search to simulate a debate with the goal of identifying a series of pixels to convince a pre-trained Judge classifier of an input example's class. While conceptually similar to advocacy learning, our work differs in motivation and methodology. In addition to considering a different learning framework: neural networks, vs. Monte-Carlo tree search, there are two key differences in problem formulation: 1) our work does not use a pre-trained Judge, and 2) our work involves Advocates, which are assigned classes, not debaters, which choose classes.

\section{Conclusion}
We have presented a novel approach to attention-based supervised classification: advocacy learning. Our approach divides a network into two subnetworks: i) a set of Advocates trained to provide arguments supporting their corresponding class, and ii) a Judge that uses these arguments to predict the true class. Over a series of experiments on three publicly available datasets, we showed that class-conditional attention can improve performance relative to standard attention, and that in some circumstances competitive training can further improve performance.

The use of a deep network for the Judge has important implications for the interpretability of the derived attention maps. If the Judge is a high-capacity nonlinear network, then the evidence it may find convincing will, by default, be uninterpretable. However, the flexibility of the proposed architecture means that work on training interpretable networks or interpreting trained networks applies \cite{ribeiro_model-agnostic_2016}. Future work could consider using these techniques to create interpretable Advocacy Nets. Extensions may also consider improving the balance between cooperation and competition by controlling the ratio of honest and deceptive updates, or the ratio of class-specific updates. Currently, the proposed architecture is limited by the one-to-one relationship between the number of classes and number of advocates, which makes training on datasets like ImageNet infeasible. Future work could examine methods to remove this linear relationship, such as training advocates that work across class hierarchies. While there are many avenues for future improvements, the experiments explored in this work suggest that competition and class-conditional representations can in some cases be in used to improve the utility of attention in classification tasks. 

\bibliographystyle{named}
\bibliography{references}

\end{document}


\maketitle
\section{Appendix}
We provide the specific architecture used for our Judge and Advocate Module on the image and MIMIC experiments. Other implementation details are found in the main paper.
\clearpage

\begin{table}
	\centering
	\caption{The Judge network for Image data. Note that we use a more complicated network for experiments involving CIFAR-10, taken from Hasanpour [2016]}
	\label{table:architecture_ji}
	\begin{tabular}{ll}  
		\toprule
		Layer & Filter \\
		\midrule
		\textbf{Conv Features} & \\
		Conv1 & 32x3x3 Convolution\\
		& 32 Channel 2d BatchNorm\\
		& ReLU \\
		Conv2 & 32x3x3 Convolution\\
		& 64 Channel 2d BatchNorm\\
		& ReLU \\
		& 2x2 Max Pool \\
		Conv3 & 64x3x3 Convolution\\
		& 64 Channel 2d BatchNorm\\
		& ReLU \\
		Conv4 & 64x3x3 Convolution\\
		& 32 Channel 2d BatchNorm\\
		& ReLU \\
		& 2x2 Max Pool \\
		\textbf{Output} & \\
		FC1 & 512 node Linear layer \\
		& 512 Channel 1d BatchNorm \\
		& ReLU \\
		& Dropout($p=0.2$)\\
		Out & Linear Output\\
		\bottomrule
	\end{tabular}
\end{table}

\begin{table}
	\centering
	\caption{The Advocate Module for Image data. Note the final convolution has a number of layers equal to the input channel size.}
	\label{table:architecture_ai}
	\begin{tabular}{ll}  
		\toprule
		Layer & Filter \\
		\midrule
		\textbf{Encoder} & \\
		Conv1 & 32x3x3 Convolution\\
		& 32 Channel 2d BatchNorm\\
		& ReLU \\
		Conv2 & 32x3x3 Convolution\\
		& 64 Channel 2d BatchNorm\\
		& ReLU \\
		& 2x2 Max Pool \\
		Conv3 & 64x3x3 Convolution\\
		& 64 Channel 2d BatchNorm\\
		& ReLU \\
		Conv4 & 64x3x3 Convolution\\
		& 32 Channel 2d BatchNorm\\
		& ReLU \\
		& 2x2 Max Pool \\
		\textbf{Decoder} & \\
		Deconv1 & 32x3x3 Stride-1 Deconvolution\\
		& 32 Channel 2d BatchNorm\\
		& ReLU \\
		Deconv2 & 16x2x2 Stride-2 Deconvolution\\
		& 16 Channel 2d BatchNorm\\
		& ReLU \\
		Deconv3 & 8x2x2 Stride-2 Deconvolution\\
		& 8 Channel 2d BatchNorm\\
		& ReLU \\
		Deconv4 & 4x5x5 Stride-1 Deconvolution\\
		& 4 Channel 2d BatchNorm\\
		& ReLU \\
		Output & 2x3x3 Convolution with Padding=1 \\ 
		& 2 channel 2d BatchNorm\\
		& Nx1x1 Convolution\\
		\bottomrule
	\end{tabular}
\end{table}

\begin{table}
	\centering
	\caption{The Judge network for MIMIC III.}
	\label{table:architecture_jM}
	\begin{tabular}{ll}  
		\toprule
		Layer & Filter \\
		\midrule
		\textbf{Conv Features} & \\
		Conv1 & 64x3 1D Convolution\\
		& 64 Channel 1D BatchNorm\\
		& ReLU \\
		& 2-width 1D Max Pool \\
		Conv2 & 64x3 1D Convolution\\
		& 64 Channel 1D BatchNorm\\
		& ReLU \\
		& 2-width 1D Max Pool \\
		FC1 & 64 node Linear layer \\
		& 64 Channel 1D BatchNorm \\
		& ReLU \\
		& Dropout($p=0.2$)\\
		Out & Linear Output\\
		\bottomrule
	\end{tabular}
\end{table}

\begin{table}
	\centering
	\caption{The Advocate Module for MIMIC. Note the final convolution has a number of layers equal to the input channel size, which for the MIMIC III benchmark is 76.}
	\label{table:architecture_aM}
	\begin{tabular}{ll}  
		\toprule
		Layer & Filter \\
		\midrule
		\textbf{Encoder} & \\
		Conv1 & 32x3 1D Convolution\\
		& 32 Channel !d BatchNorm\\
		& ReLU \\
		Conv2 & 32x3 1D Convolution\\
		& 32 Channel 1d BatchNorm\\
		& ReLU \\
		& 2x2 Max Pool \\
		Conv3 & 64x3 1D Convolution\\
		& 64 Channel 1d BatchNorm\\
		& ReLU \\
		Conv4 & 64x3 1D Convolution\\
		& 64 Channel 2d BatchNorm\\
		& ReLU \\
		& 2x2 Max Pool \\
		\textbf{Decoder} & \\
		Deconv1 & 32x3 Stride-1 1D Deconvolution\\
		& 32 Channel 1D BatchNorm\\
		& ReLU \\
		Deconv2 & 32x2 Stride-2 1D Deconvolution\\
		& 32 Channel 1D BatchNorm\\
		& ReLU \\
		Deconv3 & 64x2 Stride-2 1D Deconvolution\\
		& 64 Channel 1D BatchNorm\\
		& ReLU \\
		Deconv4 & 64x5 Stride-1 1D Deconvolution\\
		& 64 Channel 1d BatchNorm\\
		& ReLU \\
		Output & 64x3 1D Convolution with Padding=1 \\ 
		& 64 channel 1D BatchNorm\\
		& Nx1 1D Convolution\\
		\bottomrule
	\end{tabular}
\end{table}